\documentclass[10pt,twocolumn,letterpaper]{article}

\usepackage{cvpr}
\usepackage{times}
\usepackage{epsfig}
\usepackage{graphicx}
\usepackage{amsmath}
\usepackage{amssymb}

\usepackage[utf8]{inputenc}

\usepackage{algorithm,algpseudocode}
\usepackage{array,multirow}
\usepackage{subfigure}
\usepackage{float}
\usepackage{color}
\usepackage{ulem}

\graphicspath{{fig/}{images/}}

\usepackage{tabularx}
\newcolumntype{Y}{>{\centering\arraybackslash}X}

\usepackage{bm}

\newcommand{\bx}{{\mathbf{x}}}
\newcommand{\by}{{\mathbf{y}}}

\newcommand{\comment}[1]{}

\definecolor{crimson}{rgb}{0.56, 0.08, 0.14}
\definecolor{tG}{rgb}{0, 0.5, 0.25}
\ifx \submission \undefined

\else

\fi

\usepackage[pagebackref=true,breaklinks=true,letterpaper=true,colorlinks,bookmarks=false]{hyperref}

\cvprfinalcopy 

\ifcvprfinal\fi
\begin{document}

\title{Non-local RoIs for Instance Segmentation}

\author{Shou-Yao Roy Tseng\\
National Tsing Hua University\\
{\tt\small roytseng.tw@gmail.com}
\and 
Hwann-Tzong Chen\\
National Tsing Hua University\\
{\tt\small htchen@cs.nthu.edu.tw}
\and
Shao-Heng Tai\\
Umbo Computer Vision\\
{\tt\small daniel.tai@umbocv.com}
\and
Tyng-Luh Liu\\
Academia Sinica\\
{\tt\small liutyng@iis.sinica.edu.tw}
}

\maketitle

\begin{abstract}
We introduce the concept of Non-Local RoI (NL-RoI) Block as a generic and flexible module that can be seamlessly adapted into different Mask R-CNN heads for various tasks. Mask R-CNN treats RoIs (Regions of Interest) independently and performs the prediction based on individual object bounding boxes. However, the correlation between objects may provide useful information for detection and segmentation. The proposed NL-RoI Block enables each RoI to refer to all other RoIs' information, and results in a simple, low-cost but effective module. Our experimental results show that generalizations with NL-RoI Blocks can improve the performance of Mask R-CNN for instance segmentation on the Robust Vision Challenge benchmarks. 
\end{abstract}

\section{Introduction}

The current trend of deep network architectures for object detection can be categorized into two main streams: one-stage detectors and two-stage detectors. One-stage detectors perform the task of object detection in an end-to-end single-pass manner, \eg YOLO~\cite{Joseph2016YOLO, Joseph2017YOLOv2, Redmon2018YOLOv3} and SSD~\cite{Liu2016SSD, Fu2017DSSD}. On the other hand, two-stage detectors divide the task into two sub-problems that respectively focus on extracting object region proposals and classifying each of the candidate regions. Detectors such as Faster R-CNN~\cite{Ren2015Faster} and Light-Head R-CNN~\cite{Li2017lighthead} are both of this kind.

Mask R-CNN~\cite{He2017MaskRCNN} extends Faster R-CNN by adding a branch for predicting segmentation masks on each Region of Interest (RoI) in parallel with the existing branch for classification and bounding box regression. This showcases the architecture flexibility of two-stage detectors for multitasking over the one-stage counterparts. Different branches in Mask R-CNN share the same set of high-level features extracted by a deep CNN backbone network, such as ResNet~\cite{He2016ResNet}. Then, each branch attends to specific RoI via \textit{RoIAlign}, a simple and quantization-free layer that faithfully preserves spatial preciseness. Further, the proposed {\em Non-Local RoI (NL-RoI) Block} can be incorporated into Mask R-CNN to achieve better performance.

The ability to capture long-range and non-local information is a key success factor of deeper CNNs. For vanilla Mask R-CNN, the only means to acquire non-local information for each RoI is to explore the high-level features extracted by the deep backbone network. However, the high-level features are shared among all RoIs of different spatial locations, semantic categories, and branches for different tasks. Such high-level features are assumed to be general rather than specific for individual RoIs so that they are applicable to all the above varieties. Therefore, it is difficult for the same set of features to also contain the RoI-specific information. Besides, RoI features are rectangularly extracted based on their corresponding bounding box proposed by the \textit{Region Proposal Network (RPN)}. It is very likely to have multiple instances in a single bounding box when the scene is crowded. Moreover, if the instances are of the same category, it is harder for the branch network to tell apart the boundary by only referring to the local feature within an RoI. Especially for non-rigid objects, such as persons, the target object will deform in shape, and the bounding box has a higher chance to include other objects interlacing in a more complicated way.

To tackle the above concern, we introduce the idea of NL-RoI Block to better address the problem, and argue that RoI-specific non-local information can be helpful in discriminating the target instance from the others. For example, due to object co-occurrence prior in the real world, it is more probable to see cars along with pedestrians instead of refrigerators in a street scene. Besides, mutual information between instances may also be useful. Consider a scene of group dancing: People are usually posing in similar ways, and hence we can more confidently predict the pose for a dancer under partial occlusion, by referring to other dancers' poses.


Our NL-RoI Block module is inspired by the non-local operations proposed by Wang \etal~\cite{Wang2018Nonlocal}. They present the non-local operations as a family of generic building blocks for capturing long-range dependencies in different locations of data domain. The location can sit in a pixel or an audio sample for visual and acoustic data respectively. For visual data domain, the dependencies may come across space for tasks using a single static image, or space-time for tasks involving an extra time dimension such as video classification. In contrast, NL-RoIs are focusing on the long-range dependencies at \textbf{a higher level between instances} instead of just the pixel level. Specifically, our method  explicitly empowers the network to model correlations and attentions between RoIs. By taking into account all pairs of RoIs of a scene in an efficient way, the NL-RoI Block benefits from not only neighboring RoIs but also spatially separated ones.

\section{Non-local RoI}
\begin{figure}[t]
  \centering
  \includegraphics[width=0.985\linewidth]{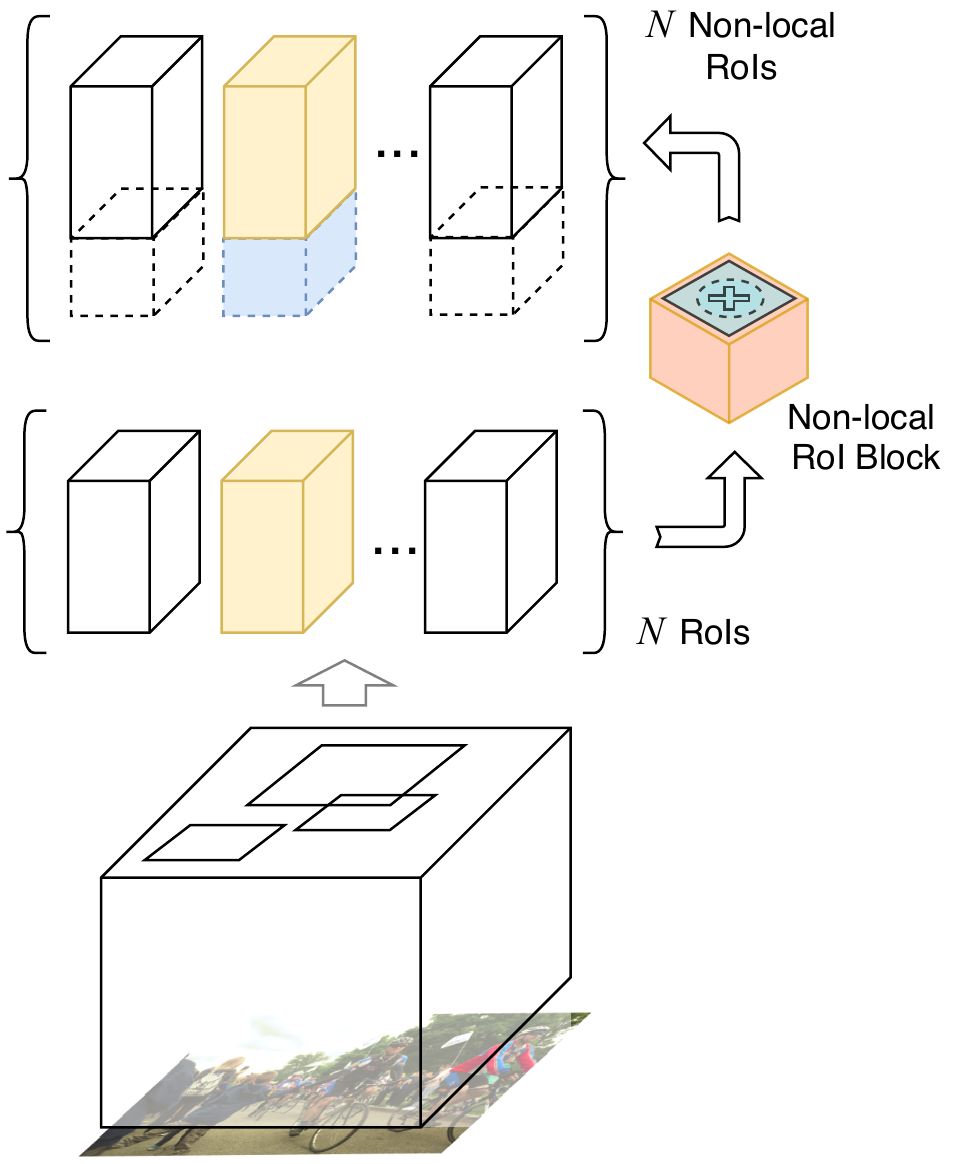} 
  \caption{Using an NL-RoI Block to extract augmented RoI-specific features.
  }
  \label{fig:NLB}
\end{figure}

We first introduce the general definition of non-local RoI operation by following the notations in \cite{Wang2018Nonlocal}. We then go on to provide a detailed implementation about the NL-RoI Block used in Robust Vision Challenge 2018. Fig.~\ref{fig:NLB} shows the basic idea about how we apply the NL-RoI Block to augment the original RoI feature blobs.

\subsection{Formulation}
Inspired by the non-local operation in \cite{Wang2018Nonlocal}, we define a generic non-local RoI operation for the use in conjunction with R-CNN based models \cite{Ross2014RCNN}:
\begin{equation}
\begin{aligned}	\by_{i} = \frac{1}{\bm{C}(X)_i} \sum_{j=1}^{N}{ f(\mathbf{x}_{i}, \bx_{j}) g(\bx_{j})} \,,
\end{aligned}
\label{equation:nonlocal_roi}
\end{equation}
\noindent
where $i$ is the index of a target RoI whose non-local information is to be computed and $j$ enumerates all the $N$ RoIs, including the target one. The input feature blob is denoted as $X=[\bx_1,\cdots,\bx_N]$ and the output feature containing non-local information is denoted by $Y=[\by_1,\cdots,\by_N]$. A pairwise function $f$ computes a scalar that reflects the correlation between the $i$th target RoI and each of the RoIs ($\forall{j} \in \{1..N\}$). The unary function $g$ maps the input feature from the $j$th RoI to another representation, which gives the operation the capacity to convert the input feature to be more specialized for non-local information. Finally, the response is normalized by a factor $\bm{C}(X)_i$.

The non-local RoI property in Eq.~(\ref{equation:nonlocal_roi}) originates from the fact that all RoIs are associated with each other in the operation. For each RoI, the non-local RoI operation computes responses based on correlations between different RoIs. Theoretically, each RoI should gradually learn to characterize a meaningful instance during training. That is, Eq.~(\ref{equation:nonlocal_roi}) enables the attention mechanism between instances. Moreover, this kind of non-local operation supports a variable input number $N$ of RoIs.

\subsection{Implementation of NL-RoI Block}
While different possible instantiations for $f$ can be chosen, Wang \etal~\cite{Wang2018Nonlocal} show, by experiments, that the non-local operations are not sensitive to specific choices. For simplicity, we just adopt the \textit{Embedded Gaussian} version of $f$:
\begin{equation}
\begin{aligned} f(\bx_{i},\bx_{j}) = e^{\phi({\bx_{i}})^{T} \psi({\bx_{j}})} \,,
\end{aligned}
\label{equation:f_embedded_gaussian}
\end{equation}
\begin{equation}
\begin{aligned}	\bm{C}(X)_i = \sum_{j=1}^N{f(\bx_{i}, \bx_{j})} \,.
\end{aligned}
\label{equation:f_normalization_factor}
\end{equation}
\noindent
Assume that we have $N$ RoIs and $D$ channels of input features, and the aligned RoI spatial size is $H \times  W$. Hence, the input feature blob $X$ has the shape of $(N, D, H, W)$. The two embedding functions $\phi$ and $\psi$ are both chosen to be a 1-by-1 2D convolution that reduces the channel dimension of the input blob. The purpose of $f$ is to calculate the correlations between $N$ RoIs, so the output of $f$ being applied to the whole input blob $X$ should be an $N$-by-$N$ matrix. The output blobs from $\phi$ and $\psi$ are reshaped to $(N, D_f \times H \times W)$. Afterward, a matrix multiplication on the reshaped outputs is performed to obtain the correlation matrix. Exponential and normalization terms are implemented by taking \textit{softmax} to the rows of the correlation matrix.

It is worth noting that this form of $f$ is essentially the same as the \textit{Self-Attention Module} in \cite{Vaswani2017Attention} for machine translation. For a given $i$, $\frac{1}{\bm{C}(X)_i} f(\bx_{i},\bx_{j})$ becomes a \textit{softmax} computation along the dimension $j$. Eq.~(\ref{equation:nonlocal_roi}) results in the self-attention form $Y =\textit{softmax}(X^{T}W_{\phi}^{T}W_{\psi}X)$ in \cite{Vaswani2017Attention}.

\begin{figure}[t]
  \centering
  \includegraphics[width=0.9\linewidth]{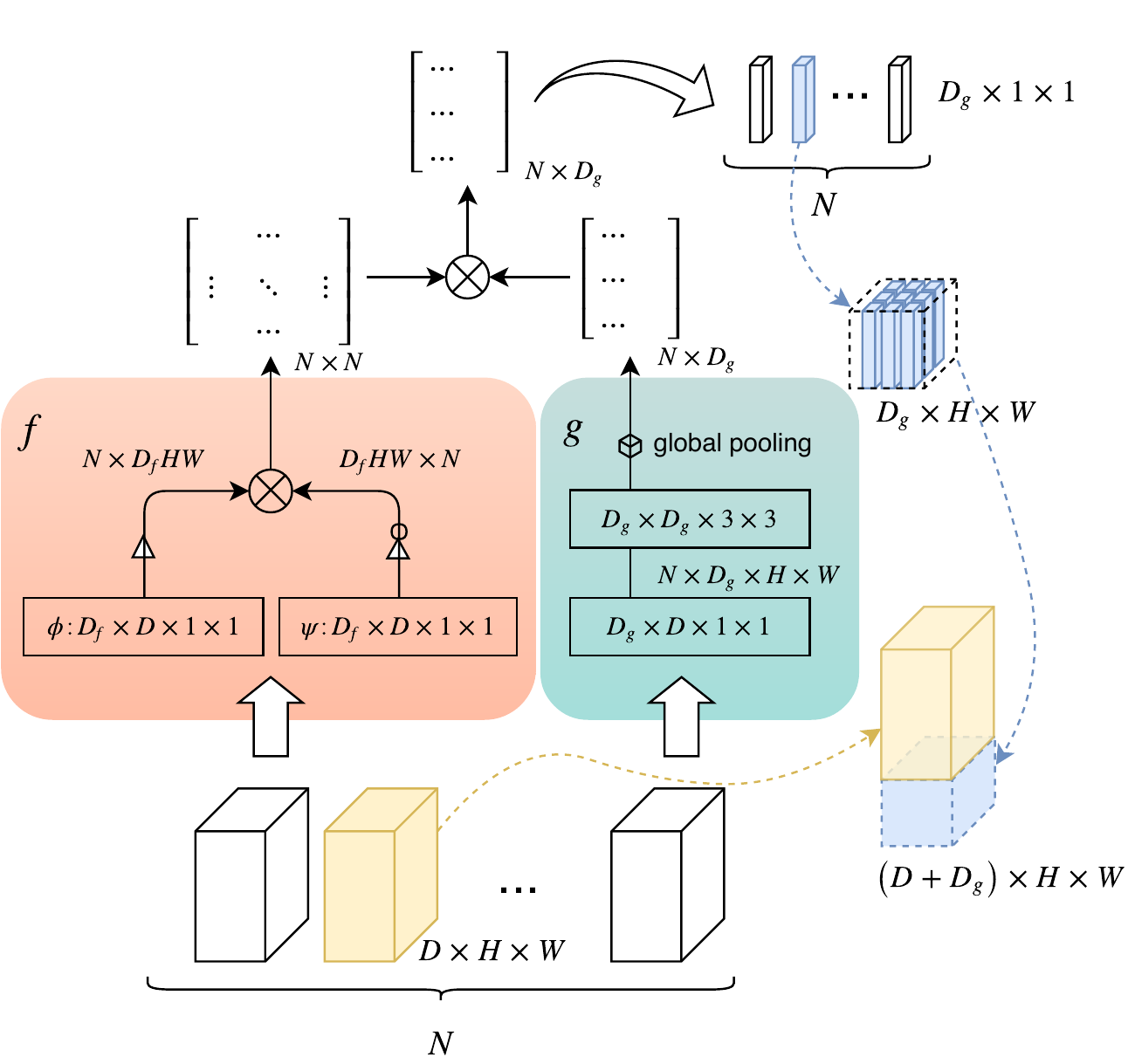}
  \caption{The detailed operations of a NL-RoI Block.}
  \label{fig:NLB_detail}
\end{figure}

The remaining part in non-local RoI operation $g$ is responsible for extracting useful non-local information from the input feature. Following the bottleneck design of \cite{He2016ResNet}, we first use a 1-by-1 convolution to reduce the channel dimension and then a 3-by-3 convolution to take in the spatial information. To further cut down memory cost, a global 2D average pooling is applied. Finally, the pooled feature blob of shape $(N, D_g, 1, 1)$ is tiled around $H, W$ spatial dimensions and is appended to the end of input blob, as showed in Fig.~\ref{fig:NLB_detail}. A ReLU activation function \cite{Nair2010ReLU} is used between the two convolution layers.

\section{Instance Segmentation Model}
Our NL-RoI Block is plugged into Mask R-CNN to perform instance segmentation. The backbone network for image feature extraction is ResNet-50 with FPN \cite{Lin2017FPN}. We replace batch normalization \cite{Ioffe2015batchnorm} by group normalization \cite{Wu2018groupnorm} for better training stability and convergence with a smaller batch size.

\paragraph{Training.} The core training datasets for our method include Cityscapes~\cite{Cordts2016Cityscapes}, Kitti Instance Segmentation~\cite{Alhaija2017BMVC}, WildDash~\cite{ZendelMHH2017WildDash}, and ScanNet~\cite{Dai2017ScanNet}. In addition, we use ADE20K~\cite{Zhou2017scene} to provide more furniture samples for training. There are 76{,}528 valid training images in total. We train for 136K iterations, starting from a learning rate of $0.02$ and reducing it to $0.01$, $1e{-3}$, $1e{-4}$ on 56K\textsuperscript{th}, 76K\textsuperscript{th}, 116K\textsuperscript{th} iteration respectively. We use a weight decay of 0.0001 and a momentum of 0.9. Pre-trained weights for corresponding Mask R-CNN architecture from Detectron~\cite{Detectron2018} are loaded during  initialization.

\paragraph{Inference.} At inference time, the input image is resized to 800 pixels on the shorter side. If the length of the longer side of resized image exceeds 1{,}333 pixels, we further resize the image to make sure the length of the longer side is 1{,}333 pixels. Soft-NMS~\cite{Bodla2017softnms} and box-voting~\cite{GidarisK2015} are also used during inference.

\vspace{2mm}
All implementations of the proposed NL-RoI Block and the related modifications are based on PyTorch deep learning framework~\cite{paszke2017automatic} and the \textit{Detectron.pytorch} GitHub Repo~\cite{Detectron.pytorch2018} of the first author, Roy Tseng. 

\section{Benchmark Results}

\begin{figure}[h]
	\centering
	\includegraphics[width=0.455\textwidth]{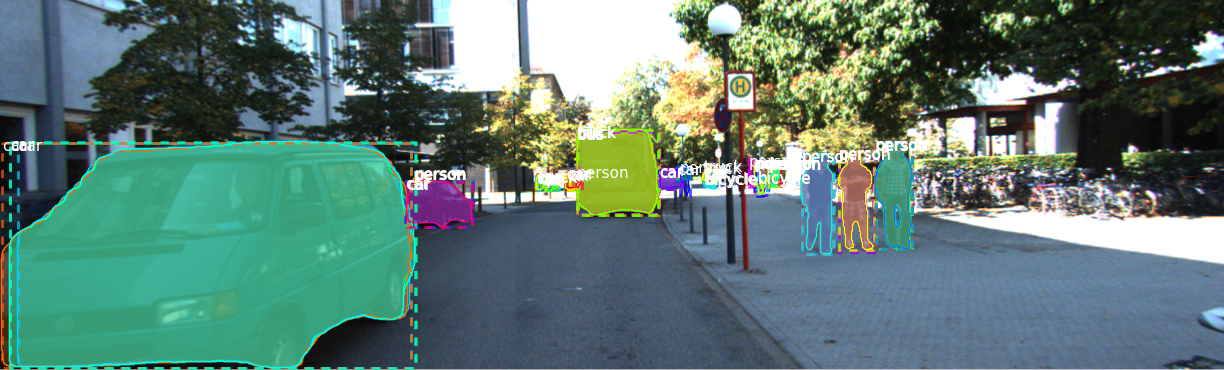}
	\includegraphics[width=0.455\textwidth]{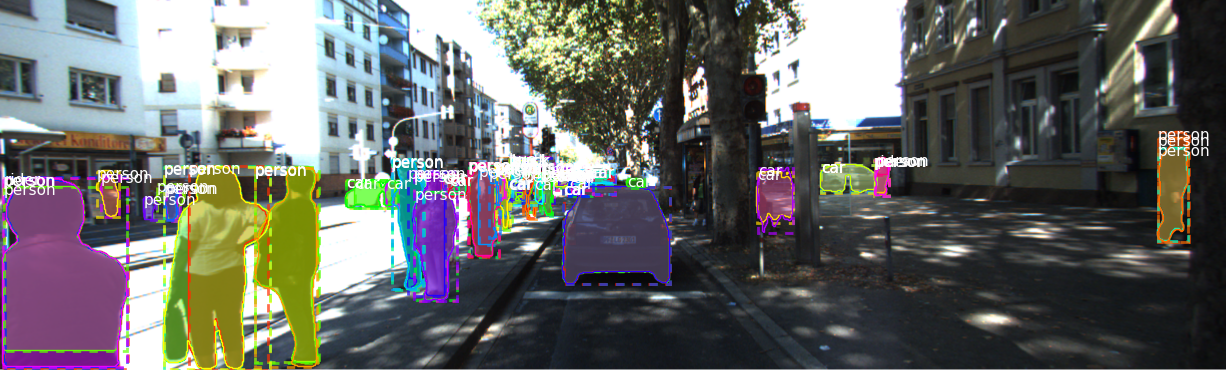}
    \caption{Instance segmentation sample results on Kitti test set.}
	\label{figs:kitti_instance_results}
\end{figure}

Table \ref{rob2018_bench} summarizes the instance segmentation benchmark results of NL-RoI on the four datasets involved in Robust Vision Challenge 2018. Fig.~\ref{figs:kitti_instance_results} shows two sample results on the Kitti test set.
\begin{table}[h]
\centering
\label{rob2018_bench}
\begin{center}
\begin{tabular}{|@{\,}c@{\,}|@{\,}c@{\,}|@{\,}c@{\,}|@{\,}c@{\,}|@{\,}c@{\,}|@{\,}c@{\,}|}
\hline
Dataset         & AP50:95  & AP50   & AP100m  & AP50m  & Neg AP \\ \hline
\sf{Kitti}      & 16.37\%  & 34.5\% &  -      & -      & -      \\ \hline
\sf{Cityscapes} & 24.0\%   & 45.8\% & 36.1\%  & 40.8\% & -      \\ \hline
\sf{WildDash}   & 19.4\%   & 34.0\% &  -      & -      & 19.7\% \\ \hline
\sf{ScanNet}    & 11\%   & -      &  -      & -      & -      \\ \hline
\end{tabular}
\end{center}
\caption{ROB2018 Instance Segmentation Benchmarks. AP: average of average precision ranging from overlap 0.5 to 0.95 in steps 0.05. AP50: average precision at overlap 0.5. AP100m/50m: average precision on objects within 100m/50m distance. Neg AP: average precision on images with visual hazards of blur, distortion, overexposure, etc.}
\end{table}

\pagebreak
{\small
\bibliographystyle{ieee}
 \bibliography{vision}
}

\end{document}